\documentclass[letterpaper, 10 pt, conference]{ieeeconf}  

\IEEEoverridecommandlockouts                              

\usepackage{graphicx}
\graphicspath{{figs/}}
\usepackage{amsmath}
\usepackage{amssymb}
\usepackage{algorithm}
\usepackage{algpseudocode}
\usepackage{times}
\usepackage{bbm}
\usepackage{array}
\usepackage{threeparttable}
\usepackage[table]{xcolor}
\usepackage{cite}
\newcommand{\figref}[1]{Fig.~\ref{figure:#1}}
\newcommand{\tabref}[1]{Table~\ref{table:#1}}

\title{\LARGE \bf
Robust Brachiation on a Life-Sized Dual-Arm Robot Using Waypoint-Guided Reinforcement Learning
}

\author{Ayumu Iwata$^{1}$, Kento Kawaharazuka$^{1}$, Keita Yoneda$^{1}$, Takahiro Hattori$^{1}$, and Kei Okada$^{1}$ 
  \thanks{$^{1}$ The authors are with the Department of Mechano-Informatics, Graduate School of Information Science and Technology, The University of Tokyo, 7-3-1 Hongo, Bunkyo-ku, Tokyo, 113-8656, Japan.
    {\texttt\small [a-iwata, kawaharazuka, yoneda, t-hattori, k-okada]@jsk.imi.i.u-tokyo.ac.jp}
  }
}

\begin{document}

\maketitle
\thispagestyle{empty}
\pagestyle{empty}

\begin{abstract}
Brachiation is a form of locomotion in which primates move primarily using their arms, enabling traversal in environments without footholds.
However, this motion requires highly coordinated whole-body movement and precise timing control for bar grasping and release.
As a result, achieving robust behavior on life-sized robotic platforms remains challenging.
In this study, we present a reinforcement learning-based method to realize brachiation on a life-sized dual-arm robot.
The core of the proposed approach is Waypoint-Guided Reinforcement Learning (WGRL), a learning framework for inducing non-linear and complex motions.
For high-difficulty tasks where imitation learning data are unavailable, WGRL guides behavior acquisition by sparsely specifying waypoints for the end-effector trajectory, while whole-body motion is generated through reinforcement learning.
In addition, by integrating the waypoint-following guidance with rewards based on task success and mechanical energy, and training in an environment designed for Sim-to-Real transfer, the proposed method achieves both forward progression and motion stability.
The acquired behavior is evaluated through Sim-to-Sim experiments under monkey-bar environments with geometric variations and hardware experiments, confirming robust brachiation including failure recovery behavior.
This study provides effective learning design guidelines for realizing arm-based locomotion on life-sized robotic hardware and expanding the traversable workspace of robots.
\end{abstract}

\section{INTRODUCTION}

Recent advances in reinforcement learning (RL) have enabled robust legged locomotion, including walking~\cite{kawaharazuka2024mevius, kawaharazuka2025mevita}, running~\cite{siekmann2021sim}, ladder climbing\cite{vogel2025robust}, and parkour~\cite{zhuang2023robot}. 
However, these behaviors remain limited to environments with footholds.
In contrast, primates achieve locomotion by suspending their bodies with their arms, even when their legs are not in contact with the ground.
Realizing such arm-based locomotion in robots would significantly expand the range of possible mobility.
A representative form of arm-based locomotion is brachiation, in which forward progression is achieved by alternately grasping and releasing overhead structures with the left and right hands.
This motion involves discontinuous contact dynamics and simultaneously requires highly coordinated whole-body movement and precise timing control.
Therefore, brachiation is a challenging motion and has attracted sustained research interest in robotics.

Most prior studies have adopted theory- or model-based approaches~\cite{kajima2003learning,fukuda2005multi,yang2019design,grama2024ricmonk}, including energy-based control and virtual holonomic constraint methods~\cite{fukuda2007design}.
While these approaches provide insight into the principles of brachiation, their reliance on accurate models limits robustness under uncertain contact conditions and disturbances.
More recently, RL approaches have been explored, including a two-stage learning approach that acquires trajectories on simplified models~\cite{reda2022learning} and a study highlighting the robustness of RL compared with optimal control on minimalistic robots~\cite{javadi2023acromonk}.
However, despite the promise of RL, most studies remain limited to simulation or small-scale systems.

\begin{figure}[t]
  \centering
  \includegraphics[width=\columnwidth]{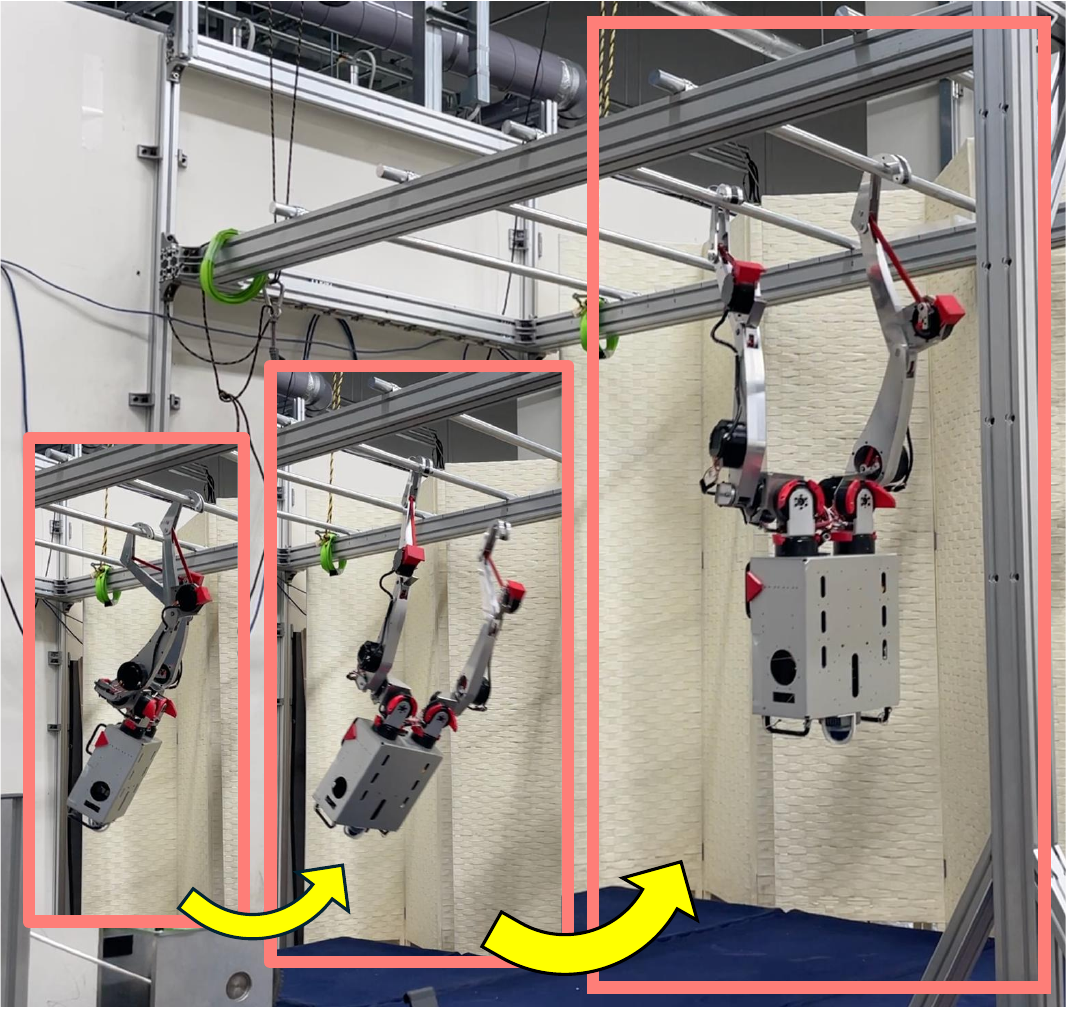}
  \vspace{-5ex}
  \caption{Brachiation motion of a life-sized dual-arm robot traversing four cylindrical bars (30 mm in diameter) arranged at 0.4 m intervals.}
  \label{figure:fig1}
  \vspace{-4ex}
\end{figure}

Compared with legged locomotion, there are two main factors that hinder the development of brachiation through RL. 
First, reward design is challenging.
Unlike walking, where design guidelines are relatively well established, systematic knowledge for brachiation remains limited.
Moreover, the complexity of brachiation makes it difficult to acquire the behavior using only simple distance- or velocity-based rewards.
In addition, high-quality reference motions for brachiation, particularly those compatible with robotic end-effector mechanisms, are scarce, making imitation learning methods such as DeepMimic~\cite{peng2018deepmimic} difficult to apply.
Second, since brachiation lacks leg support, failure directly leads to falling and episode termination, making it a high-risk task, particularly in hardware experiments.

In this study, we address the above challenges and propose a learning method to realize robust brachiation on a life-sized dual-arm robot hardware platform.
The specific contributions are as follows.

\begin{itemize}
\item We propose Waypoint-Guided Reinforcement Learning (WGRL), a learning method that induces non-linear and complex motions without imitation data.
\item We examine multiple formulations of the waypoint-following reward in WGRL and analyze how differences in its definition affect motion acquisition and learning convergence.
\item We demonstrate that integrating a success reward and a mechanical energy reward enables both forward locomotion and motion stability.
\item We validate robust brachiation with failure recovery through Sim-to-Sim evaluation under geometric variations and zero-shot Sim-to-Real transfer on life-sized robotic hardware.
\end{itemize}
\section{PROBLEM DEFINITION}

This study aims to propose a method for achieving brachiation on a life-sized dual-arm robot with hook-shaped hands in a monkey-bar environment.
The target motion consists of releasing one hand from the current bar and reaching forward to grasp the next bar ahead of the supporting hand.
Forward progression is achieved by alternately repeating this motion between the left and right hands.

\begin{figure}[t]
  \centering
  \includegraphics[width=\columnwidth]{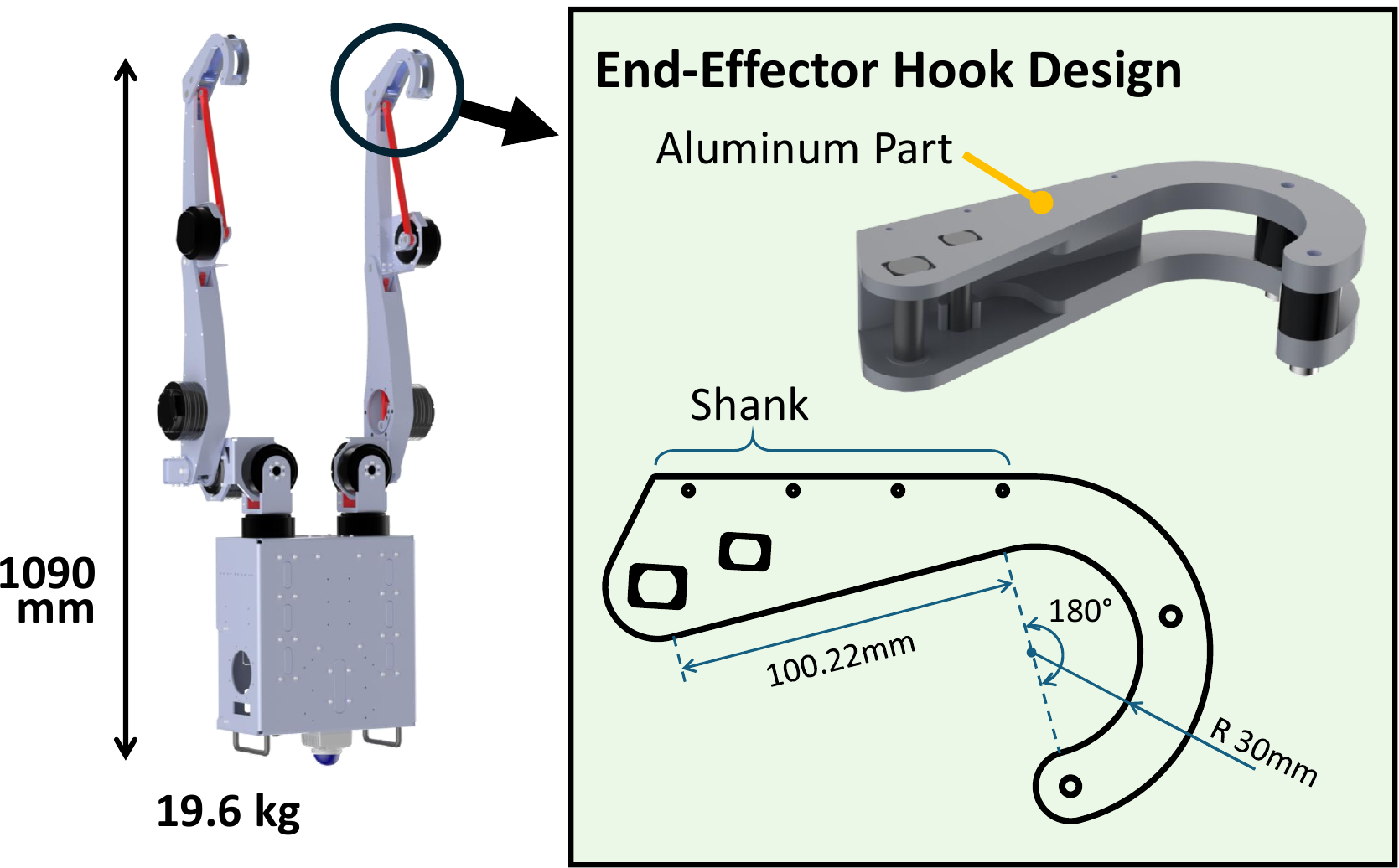}
  \vspace{-5ex}
  \caption{Mechanical design of the life-sized dual-arm robot. Left: Overview of the robot, modeled after the upper body of an adult human. Right: Hook-shaped end-effector designed for secure bar suspension and robust brachiation.}
  \label{figure:mech}
  \vspace{-4ex}
\end{figure}

\textbf{Robot}: 
We use a life-sized dual-arm robot shown in \figref{mech}.
The robot models the upper body of an adult human, with a total height of 1093 mm and a total weight of 19.6 kg.
Each arm has five degrees of freedom: shoulder roll, pitch, and yaw, elbow pitch, and wrist pitch.
This robot is developed based on the open-source hardware MEVITA~\cite{kawaharazuka2025mevita}.
The two metal-structured limbs, which provide high strength and rigidity to withstand dynamic motions, are used as arms, and the end-effectors are modified into hook-shaped hands capable of suspending from bars.
The detailed structure of the hook part is shown in the right panel of \figref{mech}.

\textbf{Monkey Bars}: 
We use a monkey-bar environment in which horizontal bars with a diameter of 30 mm and a length of 1200 mm are arranged at a constant height with intervals of 400 mm.
To promote and evaluate robustness, random variations are introduced to the position and orientation of each bar in the simulation, including the RL environment.
Specifically, variations of up to $\pm$20 mm are applied to the bar intervals, up to $\pm$40 mm in the vertical direction, and additionally up to $\pm$0.2 rad in the yaw direction relative to the robot's forward direction.
\section{BRACHIATION CONTROL WITH WGRL}

We present a control method that achieves brachiation on a life-sized dual-arm robot using RL.
As an overall procedure, the policy is first trained in a simulation environment under a reward design centered on Waypoint-Guided Reinforcement Learning (WGRL).
Through this process, a general motion pattern for brachiation is acquired.
After that, additional training is conducted under a modified reward weight configuration to obtain more refined and robust motions suitable for Sim-to-Real transfer.
The resulting policy is executed on the real robot in a zero-shot manner without additional fine-tuning.

\subsection{Policy Architecture}

An Actor-Critic architecture is adopted for the agent, and the policy is trained under an asymmetric observation setting.

\begin{table}[t]
    \centering
    \caption{Observation space used for the policies}
    \label{table:obs}
    \vspace{-2.5ex}
    \begin{tabular}{ll}
        \hline
        Name & Symbol \\
        \hline

        Base (torso link) angular velocity 
        & $\boldsymbol{\omega}_b \in \mathbb{R}^3$ \\

        Gravity direction (in torso local frame) 
        & $\boldsymbol{g}_b \in \mathbb{R}^3$ \\

        All joint positions 
        & $\boldsymbol{q} \in \mathbb{R}^{10}$ \\

        All joint velocities 
        & $\dot{\boldsymbol{q}} \in \mathbb{R}^{10}$ \\

        Motion phase indicator
        & $s_{\mathrm{phase}} \in \{0,1\}$ \\

        \hline
    \end{tabular}
    \vspace{-2ex}
\end{table}

\begin{table}[t]
    \centering
    \caption{Privileged observation space used for the critic}
    \label{table:pri_obs}
    \vspace{-2.5ex}
    \begin{tabular}{ll}
        \hline
        Name & Symbol \\
        \hline

        Joint position history
        & $\boldsymbol{q}_{t-2:t} \in \mathbb{R}^{3 \times 10}$ \\

        Joint velocity history
        & $\dot{\boldsymbol{q}}_{t-2:t} \in \mathbb{R}^{3 \times 10}$ \\

        Joint position tracking error history
        & $\boldsymbol{e}_{t-2:t} \in \mathbb{R}^{3 \times 10}$ \\

        IMU history
        & $\boldsymbol{o}^{\mathrm{IMU}}_{t-7:t} \in \mathbb{R}^{8 \times 6}$ \\

        End-effector hooking contact state
        & $\boldsymbol{c} \in \{0,1\}^{2}$ \\

        End-effector contact forces
        & $\boldsymbol{f}_{\mathrm{c}} \in \mathbb{R}^{6}$ \\

        Friction coefficients
        & $\boldsymbol{\mu} \in \mathbb{R}^{k}$ \\

        End-effector positions
        & $\boldsymbol{p}^{L,R} \in \mathbb{R}^{6}$ \\

        Five nearby bars geometry
        & $\boldsymbol{b}_{1:5} \in \mathbb{R}^{5 \times 6}$ \\

        \hline
    \end{tabular}
    \vspace{-4.5ex}
\end{table}

\textbf{Actor}: 
The actor receives as input the observations available from the hardware, as shown in \tabref{obs}.
The motion phase indicator $s_{\mathrm{phase}}$ is a command input from the controller, where $s_{\mathrm{phase}} = 0$ denotes the left-hand motion phase and $s_{\mathrm{phase}} = 1$ denotes the right-hand motion phase.
By explicitly providing motion phase information, the policy can recognize the current phase and generate appropriate actions accordingly.
The output of the actor is the target angle for each joint, which is fed into a PD controller.
The policy runs at 50 Hz, while the PD controller operates at 200 Hz.

\textbf{Critic}: 
In addition to the observations used by the actor, the critic receives privileged observations obtainable only in simulation, as shown in \tabref{pri_obs}.
The observation includes the history of the most recent three steps for joint information and eight steps for IMU information, where one step corresponds to 0.02 s, allowing the policy to capture short-term motion dynamics.
The observation of the five bars near the robot, denoted as $\boldsymbol{b}_{1:5}$, includes the center position, orientation, length, and thickness of each bar, expressed in the base link coordinate frame.
Privileged observations improve value estimation and stabilize learning.

\subsection{RL Environment Setup}

An RL environment is constructed based on Legged Gym~\cite{rudin2022learning}.
Training is conducted with 4096 parallel environments.
To improve robustness and overcome the Sim-to-Real gap, random noise is added to the observations input to the actor.
In addition, domain randomization is applied to the friction coefficients, link masses, and centers of mass.
At the beginning of each episode, the posture and motion phase are randomly selected.

\subsection{Overview of WGRL}

\begin{figure}[t]
  \centering
  \includegraphics[width=0.98\columnwidth]{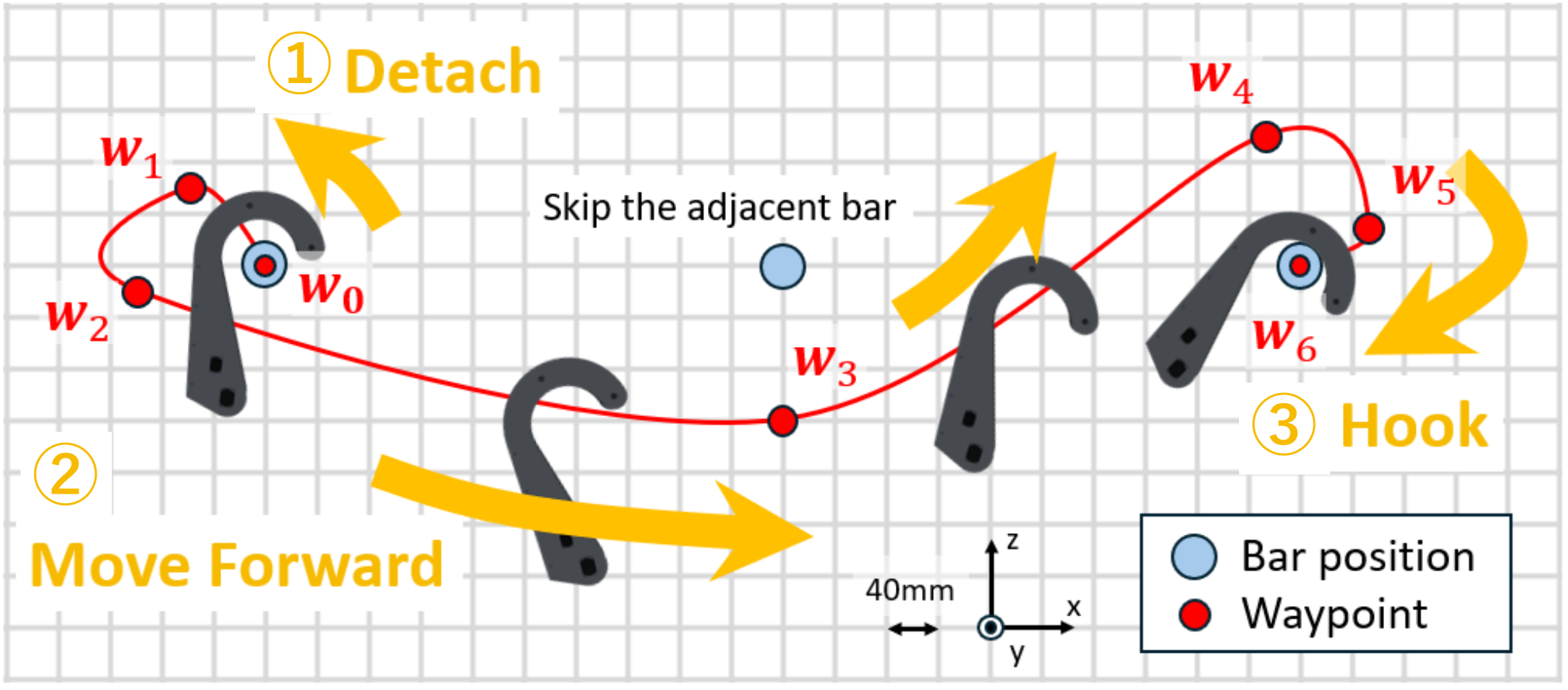}
  \vspace{-2.5ex}
  \caption{Waypoint design for inducing non-linear hooking trajectories in WGRL. The reaching hand is guided to detach, move forward beneath the bars while avoiding the support-hand bar, and hook onto the target bar from above.}
  \label{figure:waypoints}
  \vspace{-3.8ex}
\end{figure}

Waypoint-Guided Reinforcement Learning (WGRL) is a learning framework that induces non-linear and complex motions by guiding the trajectory of representative points essential for task execution through sparsely specified waypoints, while leaving whole-body motion generation to RL.
Specifically, in this study targeting brachiation, the hand position is selected as the representative point, and waypoints are predefined based on an ideal motion trajectory.
Subsequently, the policy begins exploration under a reward function that encourages the end-effector to sequentially follow these waypoints.
As learning progresses and a basic motion pattern is acquired, the waypoint guidance is gradually relaxed through a curriculum strategy by enlarging the distance threshold for reaching each waypoint.

WGRL has two key features.
First, unlike previous waypoint-based RL methods such as~\cite{mehta2024waypoint}, which use waypoints as intermediate task goals, WGRL employs waypoints only to guide early-stage exploration.
Combined with the curriculum strategy, this design enables both stable initial exploration and diverse final motions.
Second, WGRL can be applied even when full-body reference motion data for imitation learning are unavailable.

The waypoint positions specified for acquiring brachiation in this study are shown in \figref{waypoints}.
The waypoints are defined on the XZ plane (sagittal plane) and are alternately assigned to the left and right end-effectors according to the motion phase.
The first waypoint $ \{\boldsymbol{w}_1\} $ induces a motion that rotates the hand backward to detach the hook from the bar, preventing inappropriate motions that attempt to push the hand forward while still grasping the bar.
The second and third waypoints $ \{\boldsymbol{w}_2, \boldsymbol{w}_3\} $ guide the hand to pass beneath the bar and move forward, avoiding unnecessary contact with the bar located before the target.
The fourth and fifth waypoints $ \{\boldsymbol{w}_4, \boldsymbol{w}_5\} $ guide the hand to hook onto the target bar from above.
The fourth waypoint lifts the end-effector sufficiently upward, and the fifth forms a trajectory that wraps around the bar from the far side, improving the success rate of hooking.

\subsection{Reward Function Definitions}

\textbf{Reward for WGRL}: 
The reward function that encourages the end-effector to sequentially follow the waypoints is defined as a weighted sum of three reward terms, as expressed in the following equation.

\begin{equation}
    r_{\mathrm{WGRL}} = w_{\mathrm{pos}} r_{\mathrm{pos}} + w_{\mathrm{prog}} r_{\mathrm{prog}} + w_{\mathrm{succ}} r_{\mathrm{succ}}
\end{equation}

\begin{table}[t]
    \centering
    \renewcommand{\arraystretch}{1.1}
    \caption{Reward Definitions for Brachiation}
    \label{table:reward}
    \vspace{-2.5ex}
    \resizebox{\columnwidth}{!}{
    \begin{tabular}{p{2.71cm}|p{3.4cm}| c@{\hspace{3.3pt}}c}
        \hline
        \rowcolor{gray!20}
        \textbf{Name} & \textbf{Equation*} & \multicolumn{2}{l}{\textbf{Weight$^{\dagger}$}} \\
        \rowcolor{gray!20}
         &  & 1st & 2nd \\
        \hline

        WGRL (Position) &
        $r_{\mathrm{pos}}$
        & 0.64 & 0.8 \\

        WGRL (Progress) &
        $r_{\mathrm{prog}}$
        & 4.0 & 5.0 \\

        WGRL (Success) &
        $r_{\mathrm{succ}}$
        & 10 & -- \\

        Support-Hand Grasp &
        $\mathbb{I}\{\text{support-hand grasp}\}$
        & 0.5 & 0.5 \\

        Bars Reached &
        $N_{\mathrm{bar}}$
        & 2 & 0.3 \\

        Mechanical Energy &
        $- (e - e_{\mathrm{ref}})^2 + 0.5$ 
        & -- & 0.8\\

        Hook Orientation &
        $\sum (\phi_{\mathrm{hook}}^2 + \psi_{\mathrm{hook}}^2)$
        & -0.2 & -0.2 \\

        Hook Lin. Vel. &
        $\sum \|\boldsymbol{v}_{\mathrm{hook}}\|^2$
        & -0.1 & -1 \\

        Base Position &
        $y_{\mathrm{base}}^2$
        & -5e-2 & -0.1 \\

        Base Orientation &
        $\theta_{\mathrm{base}}^2$
        & -5e-2 & -0.1 \\

        Base Lin. Vel. &
        $\|\boldsymbol{v}_{\mathrm{base}}\|^2$
        & -3e-2 & -5e-2 \\

        Base Ang. Vel. &
        $\|\boldsymbol{\omega}_{\mathrm{base}}\|^2$
        & -1e-3 & -1e-2 \\

        Joint Torques &
        $\|\boldsymbol{\tau}\|^2$
        & -1e-6 & -1e-4 \\

        Joint Acceleration &
        $\|\boldsymbol{\ddot{q}}\|^2$
        & -3e-7 & -3e-7 \\

        Joint Pos. Limit &
        $|\boldsymbol{q} - \mathrm{clip}(\boldsymbol{q}, \boldsymbol{q}_{\min}, \boldsymbol{q}_{\max})|$
        & -100 & -100 \\

        Joint Vel. Limits &
        $\mathrm{clip}\bigl( |\boldsymbol{\dot{q}}| - 0.9 \, \boldsymbol{\dot{q}}_{\max}, \, 0, \, 1 \bigr)$
        & -1e-2 & -5e-2 \\

        Joint Torque Limit &
        $\mathrm{clip}\bigl( |\boldsymbol{\tau}| - 0.9 \, \boldsymbol{\tau}_{\max}, \, 0, \, \infty \bigr)$
        & -1e-2 & -5e-2 \\

        Action Rate &
        $\|\boldsymbol{a}_{t-1} - \boldsymbol{a}_{t}\|^2$
        & -1e-2 & -3e-2 \\

        Forearm Col. &
        $\sum \mathbb{I}\{\text{forearm collision}\}$
        & -20 & -20 \\

        Reaching-Hand Col. &
        $\mathbb{I}\{\text{reaching-hand collision}\}$
        & -0.5 & -0.5 \\

        Termination &
        $\mathbb{I}\{\text{end in failure}\}$
        & -70 & -70 \\
        
        \hline
    \end{tabular}
    }

    \vspace{0.5ex}
    \parbox{\columnwidth}{\footnotesize
    * $\mathbb{I}\{\dots\}$ denotes an indicator function that returns 1 if the condition inside the braces is true, and 0 otherwise. $N_{\mathrm{bar}}$ is the number of bars reached since the start of the episode. $\sum$ indicates the aggregation of the values for the left and right hands.\\
    $\dagger$ The 1st weights are used during the initial 4000 training iterations, whereas the 2nd weights are used during the subsequent 9000 iterations.}
    \vspace{-4ex}
\end{table}

\begin{itemize}
\setlength{\itemsep}{0.8em}

\item $r_{\mathrm{pos}}$ ---  a reward based on the current end-effector position.
The specific functional form of $\operatorname{f}(\cdot)$ is introduced and analyzed in the ablation study in Section~\ref{sec:wgrl_exp}.

\begin{align}
    r_{\mathrm{pos}} = \operatorname{f}(\cos\theta, \tilde{\ell})
\end{align}

Specifically, $\theta$ is defined as the angle between two vectors: 
(i) the vector from the current end-effector position to the target, and 
(ii) the vector from the previous waypoint to the target waypoint.
The cosine similarity term $\cos\theta$ is used to encourage the end-effector to approach the target from the direction of the previous waypoint.
The normalized distance term $\tilde{\ell}$ represents the current distance to the target waypoint normalized by the distance between the previous waypoint and the target waypoint.

\begin{align}
\cos\theta &= 
\frac{
(\tilde{\boldsymbol{p}}_{\mathrm{ref}} - \tilde{\boldsymbol{p}})^\top
(\tilde{\boldsymbol{p}}_{\mathrm{ref}} - \tilde{\boldsymbol{p}}_{\mathrm{prev}})
}{
\| \tilde{\boldsymbol{p}}_{\mathrm{ref}} - \tilde{\boldsymbol{p}} \|
\| \tilde{\boldsymbol{p}}_{\mathrm{ref}} - \tilde{\boldsymbol{p}}_{\mathrm{prev}} \|
}
\end{align}
\begin{align}
\tilde{\ell} &=
\frac{
\| \tilde{\boldsymbol{p}}_{\mathrm{ref}} - \tilde{\boldsymbol{p}} \|
}{
\| \tilde{\boldsymbol{p}}_{\mathrm{ref}} - \tilde{\boldsymbol{p}}_{\mathrm{prev}} \|
}
\end{align}

Here, $\boldsymbol{p}$ denotes the current end-effector position, 
$\boldsymbol{p}_{\mathrm{ref}}$ denotes the current target waypoint position, and 
$\boldsymbol{p}_{\mathrm{prev}}$ denotes the previous waypoint position.
The tilde symbol denotes projection onto the sagittal plane: $\tilde{\boldsymbol{p}} = [p_x, p_z]^\top$, where the y-axis component is omitted.

\item $ r_{\mathrm{prog}} $ --- a reward based on the decrease in the normalized distance, which encourages progress toward the target waypoint over time.

\begin{equation}
    r_{\mathrm{prog}} = \mathrm{clip} \left(\tilde{\ell}_{t-1} - \tilde{\ell}_{t}, \,\, -0.1, \,\, 0.1 \right)
\end{equation}

\item $ r_{\mathrm{succ}} $ --- a binary reward that assigns 1 when the end-effector reaches the current target waypoint and the target is updated to the next waypoint, and assigns 0 otherwise.

\end{itemize}

The target update is performed based on the comparison between the distance and the threshold $ \tilde{\ell}_{\mathrm{th}} $, as shown in the following equation.

\begin{align}
    (\boldsymbol{p}_{\mathrm{prev}}, \, \boldsymbol{p}_{\mathrm{ref}})
    &=
    (\boldsymbol{w}_{i-1}, \, \boldsymbol{w}_{i})
    \;\xrightarrow{\ \tilde{\ell} < \tilde{\ell}_{\mathrm{th}}\ }\;
    (\boldsymbol{w}_{i},\,\boldsymbol{w}_{i+1}) \notag \\
    &\quad \text{where } i \in \{1,\dots,5\}
\end{align}

The threshold is initially set to $ \tilde{\ell}_{\mathrm{th}} = 0.03 $.
As learning progresses, the threshold for each waypoint is gradually increased when the number of agents that have passed it exceeds those that have not yet reached it across the 4096 parallel environments.
This creates a curriculum structure in which strong guidance is applied in the early stage and gradually relaxed as learning proceeds.

\textbf{Other Rewards}: 
%
All reward definitions used in training are shown in \tabref{reward}.
Among them, the most important terms are the success reward (Bars Reached) and the mechanical energy reward (Mechanical Energy).
The bar-reaching success reward is given in proportion to the number of bars reached since the start of the episode.
To ensure that forward locomotion is prioritized, the weight of this reward is set to dominate the total reward.
The mechanical energy reward encourages the base-link energy to approach a prescribed target value.
The mechanical energy is defined as the sum of the kinetic energy computed from the linear velocity and the potential energy calculated based on the relative height from the bar.
The target energy is set to the value observed in stable dual-hand suspension, $e_{\mathrm{ref}} = -72\,\mathrm{J}$, to encourage stabilization after grasping and suppress excessive swing.
By combining the success reward that promotes forward progression with the mechanical energy reward that prevents excessive energy growth, both locomotion achievement and motion stability are attained.

\subsection{Training Procedure}

Proximal Policy Optimization (PPO)~\cite{schulman2017proximal} is adopted as the RL algorithm.
First, to acquire the initial motion, the weight of the WGRL reward is set large and training is performed for 4000 iterations.
After that, to refine the motion, additional training for 9000 iterations is conducted under a reward setting in which the bar-reaching success and mechanical energy terms dominate.
The specific values of the reward weights used during training are shown in \tabref{reward}.
\section{WGRL Training Experiment}
\label{sec:wgrl_exp}

\begin{figure}[t]
  \centering
  \includegraphics[width=\columnwidth]{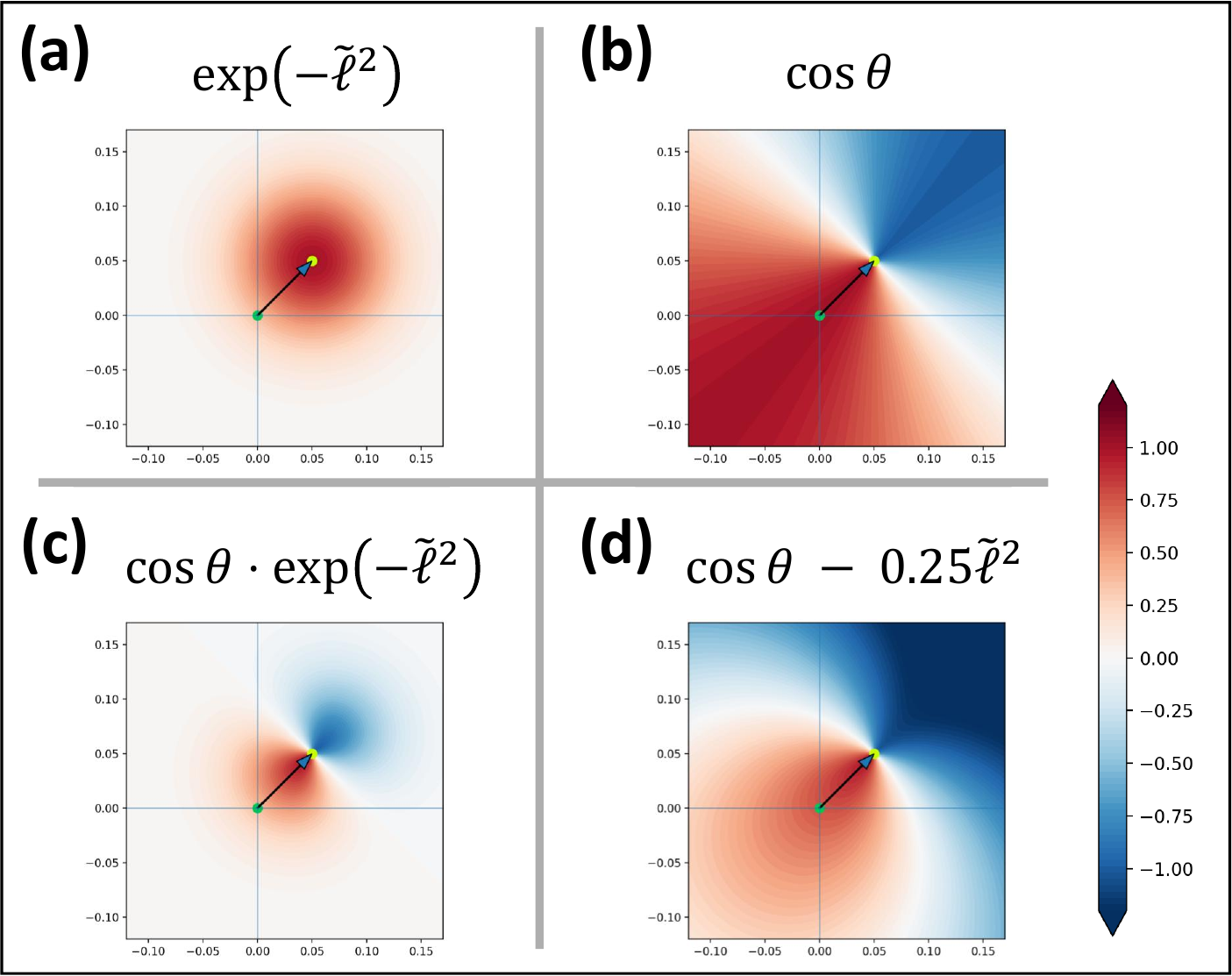}
  \vspace{-5ex}
  \caption{Reward field visualization for different definitions of $r_{\mathrm{pos}}$. (a) distance-only, (b) direction-only, (c) multiplicative distance–direction coupling, and (d) additive distance–direction formulation. The formulation (c) and (d) encourage the end-effector to approach the target from the direction of the previous waypoint.}
  \label{figure:fields}
  \vspace{-2ex}
\end{figure}

\begin{figure}[t]
  \centering
  \includegraphics[width=\columnwidth]{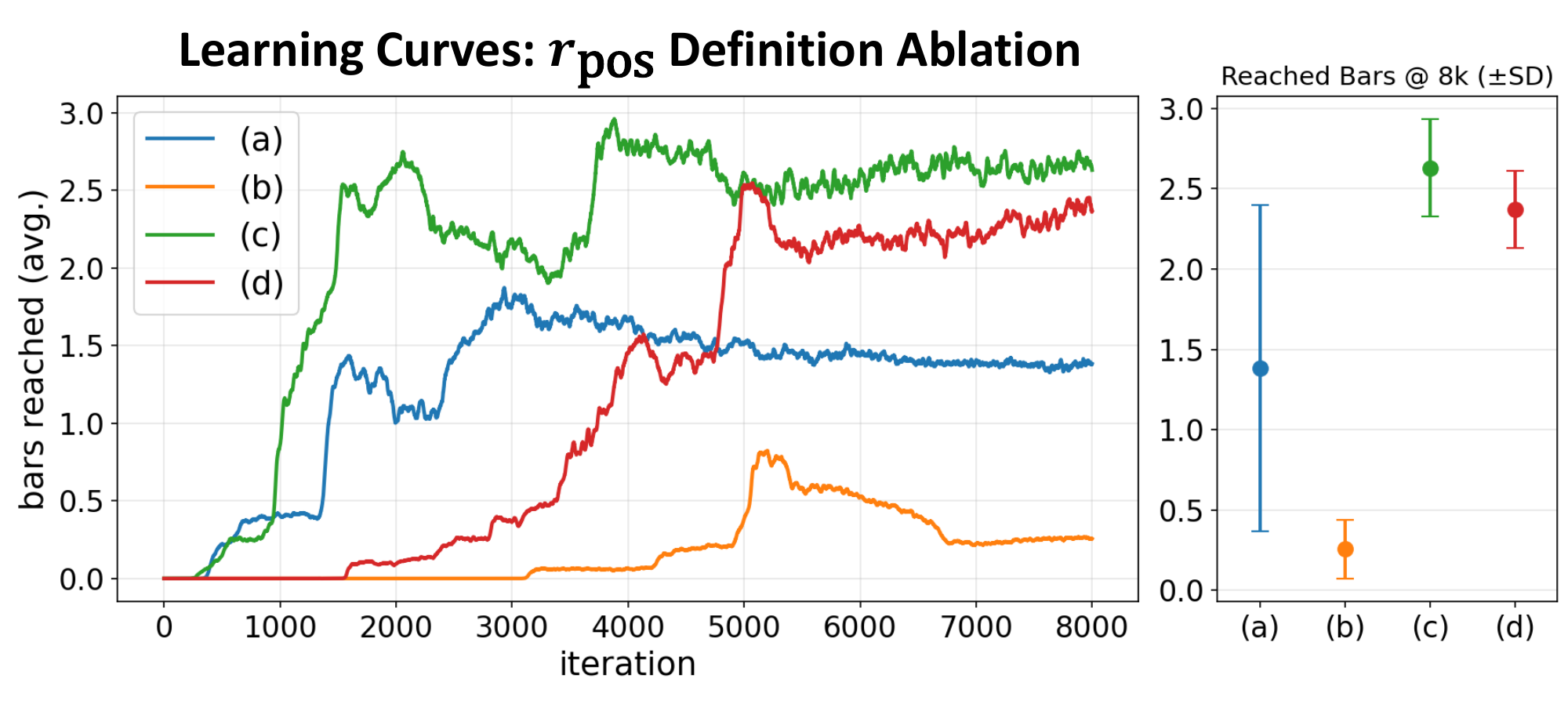}
  \vspace{-6ex}
  \caption{Learning curves for $r_{\mathrm{pos}}$ definition ablation. The vertical axis plots the number of bars reached since the start of each episode, averaged over all parallel environments. Each curve represents the mean of three independent training runs. The right panel shows the mean and standard deviation at 8000 iterations. Reward formulations incorporating both distance and directional alignment yield higher performance and more consistent convergence result.}
  \label{figure:comp2}
  \vspace{-4ex}
\end{figure}

To analyze how the definition of $r_{\mathrm{pos}} = \operatorname{f}\,(\cos\theta, \tilde{\ell})$ affects the efficiency of motion acquisition, four reward field formulations shown in \figref{fields} were compared.
All other reward terms and training conditions were kept identical, and training was conducted three times for each configuration.
The results are shown in \figref{comp2}.
The left panel shows the mean learning curves of the number of bars reached since the start of each episode, averaged across all parallel environments, while the right panel shows the mean and standard deviation at 8000 iterations.

For the distance-only definition shown in \figref{fields}(a), a high degree of exploration freedom was observed, and forward motion was occasionally acquired in the early stage of learning.
The evaluation metric exceeded 0.5 after an average of 1368 iterations, indicating a relatively fast initial rise of the learning curve.
However, without directional guidance, the policy often converged to approaching the target bar from below.
Such trajectories are physically incapable of hooking onto the bar, causing learning to stagnate.
Consequently, the standard deviation at 8000 iterations reached 1.016, indicating large performance variation and unstable learning.
For the direction-only definition based on cosine similarity shown in \figref{fields}(b), the reward lacked a clear incentive to approach the target waypoint.
As a result, the average number of bars reached at 8000 iterations was 0.257, which was the lowest performance among the four definitions.

The distance–direction combined formulations shown in \figref{fields}(c) and (d) both demonstrated stable learning convergence and high motion acquisition performance.
In particular, formulation (c) achieved the highest performance, with the evaluation metric exceeding 0.5 after an average of 954 iterations and reaching 2.6 bars at 8000 iterations.
This is attributed to its narrow positive reward region, which strongly restricts the exploration space.
On the other hand, formulation (d) includes a $-\tilde{\ell}^{2}$ term that introduces a continuous penalty increasing with the distance from the target, enabling stable approach guidance from a wide range of positions.
In this study, although formulation (c) also demonstrated strong performance, formulation (d), $r_{\mathrm{pos}} = \cos \theta - 0.25\,{\tilde{\ell}}^2$, was adopted for the subsequent experiments because of the advantage described above.

\begin{figure}[t]
  \centering
  \includegraphics[width=0.9\columnwidth]{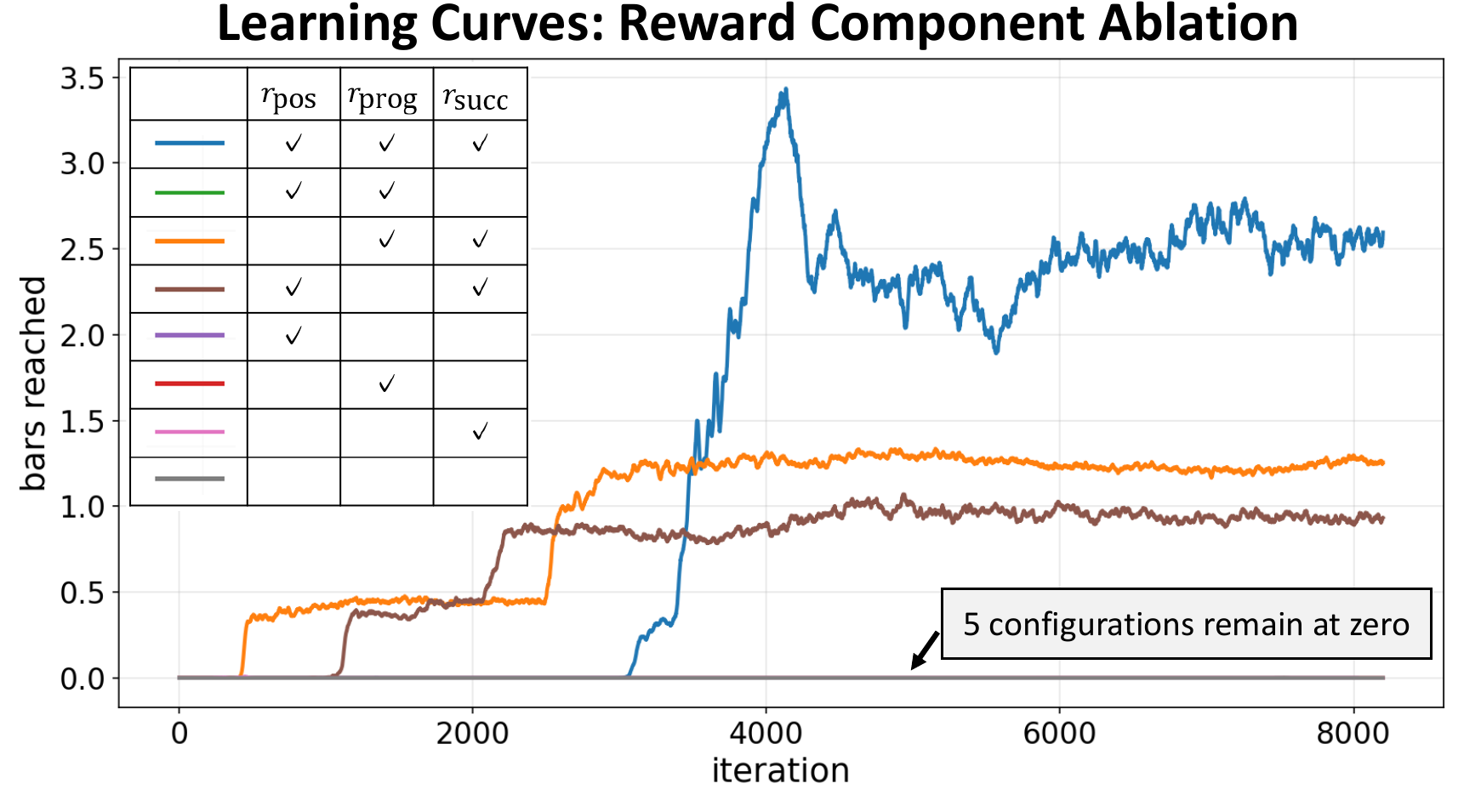}
  \vspace{-3ex}
  \caption{Learning curves for reward component ablation in WGRL. The horizontal axis indicates training iterations, and the vertical axis shows the number of bars reached since the start of each episode, averaged over all parallel environments. Brachiation behavior emerges only when all three reward components are combined, or when paired with the success reward.}
  \label{figure:comp1}
  \vspace{-4ex}
\end{figure}

Another ablation study was conducted to verify the effectiveness of the reward design in Eq.~(1) of WGRL.
Specifically, the roles of the three reward terms were analyzed by training all combinations of their presence or absence, resulting in eight configurations.
The evaluation metric was the same as in the previous experiment, namely the average number of bars reached since the start of each episode across all parallel environments.
The results are shown in \figref{comp1}.
The basic motion, which consists of releasing the hand and grasping the forward bar, was acquired only when all three reward terms were included, or when the combinations $\{r_{\mathrm{pos}}, r_{\mathrm{succ}}\}$ and $\{r_{\mathrm{prog}}, r_{\mathrm{succ}}\}$ were used.

When the waypoint-reaching success reward $r_{\mathrm{succ}}$ was excluded, learning did not progress, indicating that the success reward is indispensable as a clear incentive for waypoint traversal. 
However, when the sparse reward $r_{\mathrm{succ}}$ was used alone, learning also failed to progress. 
This result indicates that it must be combined with reward terms providing continuous feedback, such as $r_{\mathrm{pos}}$ or $r_{\mathrm{prog}}$.

When the combination $\{r_{\mathrm{pos}}, r_{\mathrm{succ}}\}$ was used, $r_{\mathrm{pos}}$ encouraged the end-effector to approach the target waypoint. 
However, since this term behaves as a potential-based reward, motion tended to stagnate in regions where the gradient was small. 
When the combination $\{r_{\mathrm{prog}}, r_{\mathrm{succ}}\}$ was used, $r_{\mathrm{prog}}$ promoted motion toward the target direction by evaluating the decrease in distance over time. 
However, because rewards are given even when the end-effector is still far from the target, sufficient approach to the waypoint is not always guaranteed.
As a result, learning converged at relatively low performance levels, with the average number of bars reached being 0.9 for $\{r_{\mathrm{pos}}, r_{\mathrm{succ}}\}$ and 1.3 for $\{r_{\mathrm{prog}}, r_{\mathrm{succ}}\}$.

In contrast, when all three reward terms were included, the complementary effects of $r_{\mathrm{succ}}$ as a motivation for reaching, $r_{\mathrm{pos}}$ as potential guidance, and $r_{\mathrm{prog}}$ as a temporal progress reward enabled stable learning and successful acquisition of brachiation.
\section{BRACHIATION RESULT AND DISCUSSION}

\subsection{Sim-to-Sim}

\begin{figure*}[t]
 \begin{center}
  \includegraphics[width=2.0\columnwidth]{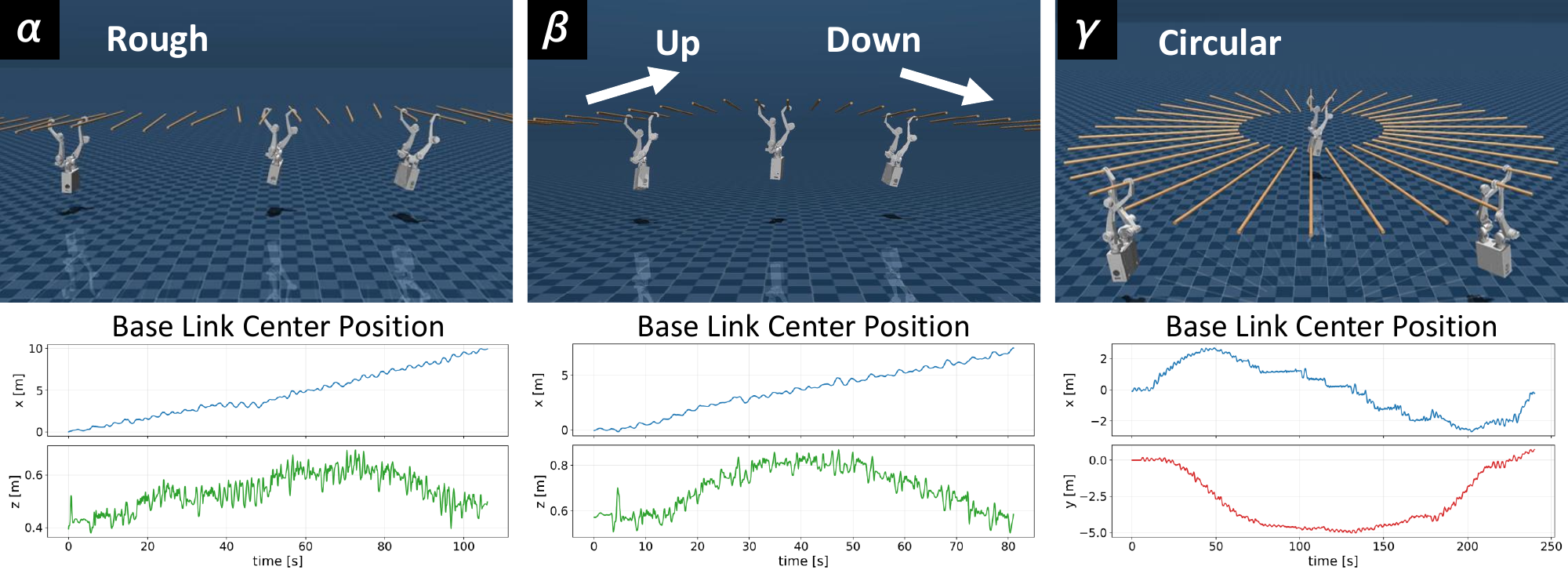}
  \vspace{-2ex}
  \caption{Sim-to-Sim evaluation in MuJoCo under diverse monkey bars configurations. The learned policy is tested in three unseen environments: ($\alpha$) geometrically perturbed bars, ($\beta$) a 5° inclined up–down course, and ($\gamma$) a circular course. The lower plots show the base link center position over time, demonstrating sustained forward progression despite environmental variations.}
  \label{figure:mujoco}
 \end{center}
 \vspace{-5ex}
\end{figure*}

To verify the robustness of the trained policy, Sim-to-Sim evaluation was conducted in a MuJoCo~\cite{todorov2012mujoco} environment.
In a monkey-bar environment with bar intervals of 400 mm, ten trials were conducted, and in all trials continuous traversal of more than 30 bars was achieved.
The robot detached one hook by swinging backward, moved it forward, and wrapped around the target bar from above to grasp it.
By alternately repeating detachment and grasping with the left and right hands, forward progression through brachiation was realized.

Next, experiments were conducted in an environment where the bars had geometric variations of up to $\pm$40 mm in the vertical direction and up to $\pm$0.2 rad in yaw relative to the robot’s forward direction.
As shown in \figref{mujoco} $\alpha$, continuous forward progression was maintained even under these perturbed conditions.
Occasionally, cases were observed in which the hook failed to catch the bar during the reaching motion.
However, the robot was able to recover and attempt to grasp the bar again.
Specifically, the recovery behavior consisted of flexion and extension of the elbow joint of the support-side arm immediately after the failure.
This motion absorbed the disturbance of balance, prevented a fall, and generated a re-swing of the reaching arm.
As a result, the robot was able to repeatedly retry the grasp and continue forward progression.

Furthermore, validation was conducted in environments completely different from the training environment, namely an up-and-down inclined course with a slope of 5 deg (\figref{mujoco} $\beta$) and a circular course composed of 36 bars (\figref{mujoco} $\gamma$).
The time-series variation of the base link position shown in \figref{mujoco} confirms that the robot adapted to previously unseen bar configurations and maintained forward progression.
In the inclined course, 28 out of 30 reaching motions were successful, corresponding to 93.3\%, and in the circular course, 91 out of 100 were successful, corresponding to 91\%.
Failure recovery behavior similar to that observed under uneven conditions was also confirmed, and grasping was often eventually achieved after multiple reattempts.

\textbf{Discussion}: 
In this study, brachiation was successfully achieved in environments with geometric variations and in unseen inclined and circular courses, even though the agent’s observations did not include explicit bar position information or visual perception.
This result suggests that the proposed method can generate robust motions that do not rely on precise estimation of the environment geometry.
Two main factors contribute to this robustness.
The first is the design of the hook mechanism.
The hook arc diameter (60 mm) provides sufficient clearance relative to the 30 mm bar, enabling passive tolerance to positional errors.
The second factor is the acquired hooking trajectory characterized by a large wrap-around motion from above.
Because of this motion, even when the bar is displaced from its expected position, the straight shank of the hook can contact the bar and slide into engagement.

The hooking trajectory can be interpreted as having been acquired through WGRL.
In the early stage of learning, the waypoint-following reward constrains the exploration toward a trajectory that wraps around the bar from above.
Afterward, even when the guidance is weakened through the curriculum strategy, the large wrap-around motion is considered to have been retained due to its stable motion performance.
These results indicate that the proposed method effectively induces non-linear and complex motion trajectories and enables stable whole-body motion based on such trajectories.

The acquisition of recovery behavior after failing to grasp the target bar is strongly influenced by the reward design.
In this study, a large penalty is imposed for falling, while a large reward proportional to successful bar grasping is provided.
Under this reward structure, the agent is encouraged to learn control strategies that avoid episode termination and maximize future success rewards through learning.
As a result, behaviors that recover balance immediately after failure to prevent falling and repeatedly attempt bar grasping emerged spontaneously.
In fact, when the weight of the success reward was reduced, the frequency of reattempts after failure significantly decreased.
This result supports the interpretation that failure recovery behavior is induced by the reward structure.
In addition, the mechanical design also enhances recoverability.
The hook arc was designed with a central angle of 180 deg, which makes the structure resistant to detachment even when whole-body motion is disturbed.
When the central angle was set to less than 180 deg in simulation, immediate detachment upon failure was frequently observed, making recovery behavior difficult to establish.

Therefore, the robust brachiation motion demonstrated in this study is enabled by both the reward design, including WGRL, and the passive mechanical role of the hook.

\begin{figure*}[t]
 \begin{center}
  \includegraphics[width=2.0\columnwidth]{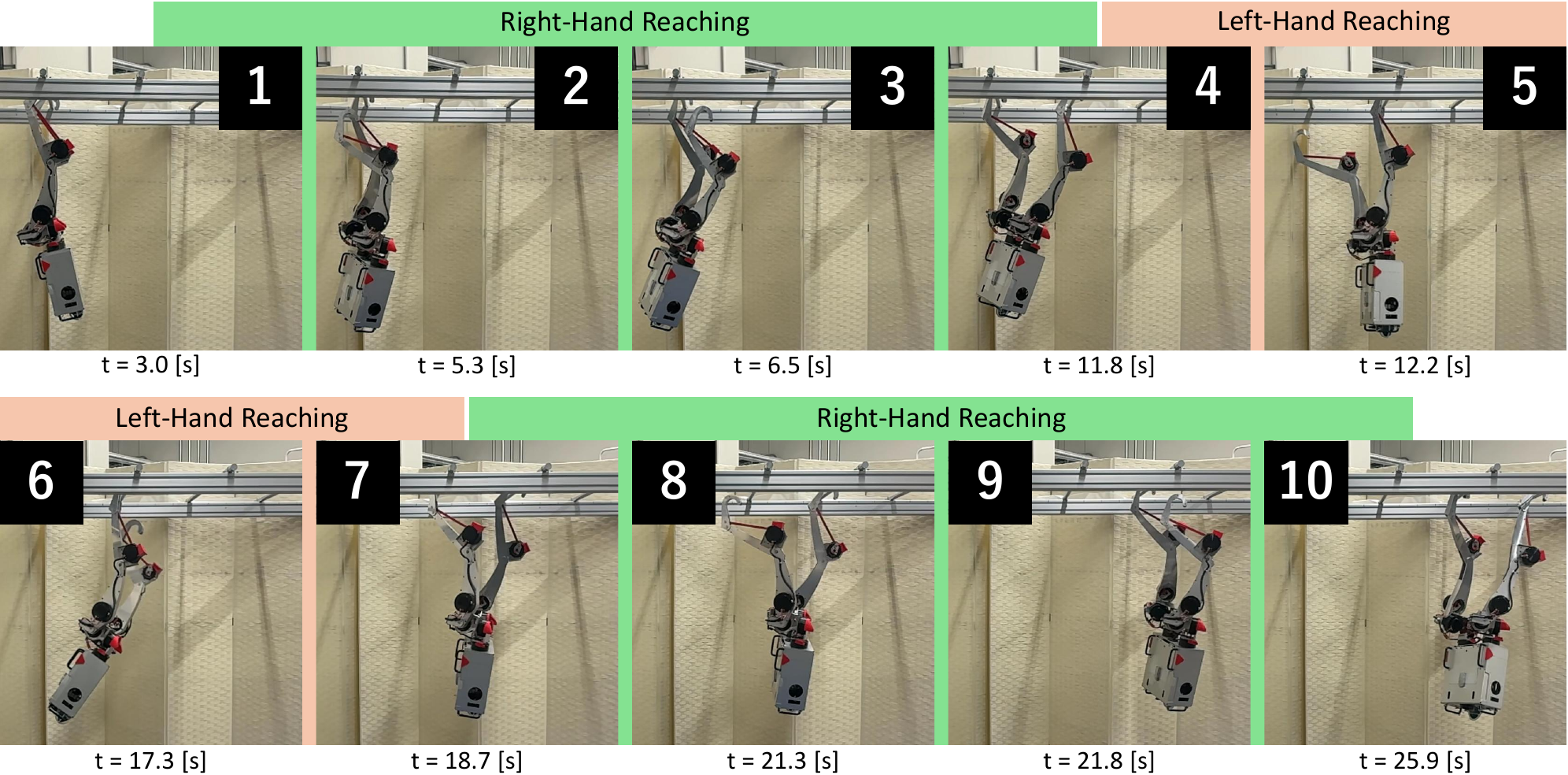}
  \vspace{-2ex}
  \caption{Zero-shot Sim-to-Real brachiation on the life-sized dual-arm robot. Sequential snapshots show alternating reaching and hooking phases across four bars spaced at 400 mm intervals.}
  \label{figure:snap}
  \vspace{-5ex}
 \end{center}
\end{figure*}

\begin{figure}[t]
  \centering
  \includegraphics[width=\columnwidth]{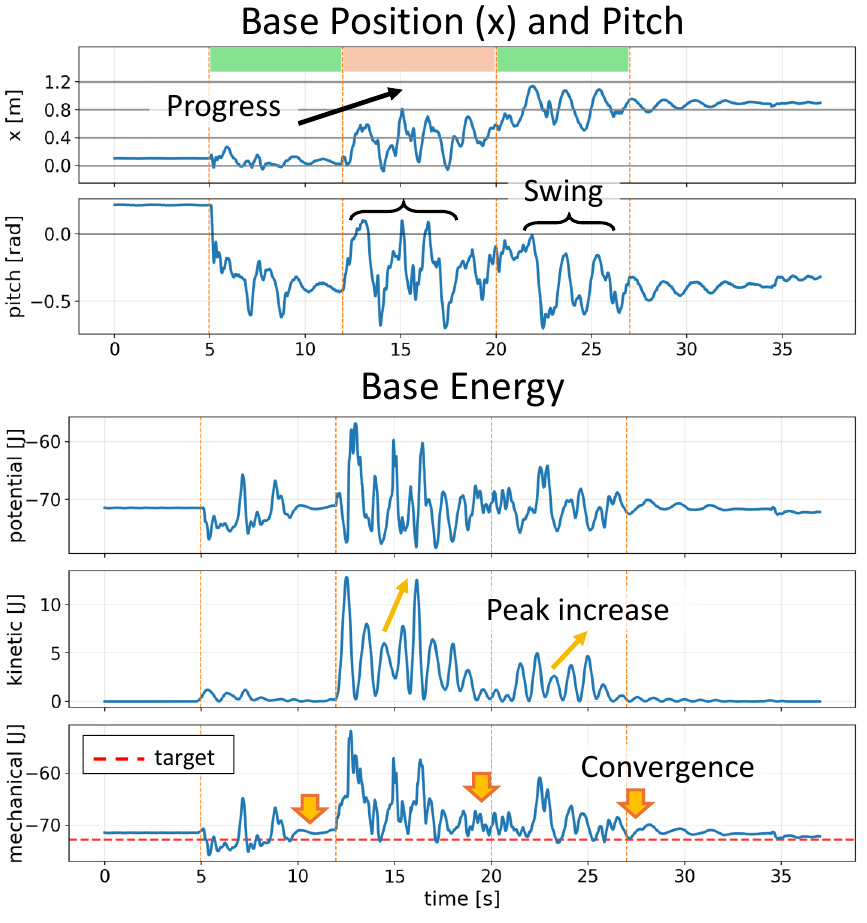}
  \vspace{-5ex}
  \caption{Hardware experiment logs during zero-shot Sim-to-Real brachiation. The top plot shows the forward base link position and pitch angle, indicating forward progression through repeated swinging motions. The bottom plot presents the base energy. During hooking failures, swing momentum increases, resulting in progressively higher kinetic energy peaks that enable reattempts. Once stable dual-hand support is established, the mechanical energy decays and converges toward the target value.}
  \label{figure:graph}
  \vspace{-5ex}
\end{figure}

\subsection{Hardware Experiment}

The policy trained in simulation was applied to the hardware robot in a zero-shot manner.
The experiment was conducted in a monkey-bar environment consisting of four cylindrical bars arranged at intervals of 400 mm.
Starting from an initial state in which both hands supported the same bar, 3 reaching phases (right, left, and right) were executed.
Although the first hooking attempt failed in each reaching phase, recovery behavior similar to that observed in the Sim-to-Sim experiments consistently emerged, enabling successful reattempts.
As a result, the reaching motion was ultimately completed in all 3 phases, and the robot traversed the entire course without falling.

\figref{snap} shows snapshots of the motion during the experiment.
\figref{graph} shows the time-series transitions of the base link position, orientation, and energy.
These values were estimated from IMU and joint angle logs using forward kinematics.
Velocity information was obtained through differentiation followed by low-pass filtering.
From the transitions of the base link's $x$ position and pitch angle, it can be confirmed that forward progression was achieved through repeated swings to attempt grasping during single-hand support.
Regarding the energy transition, during the swing phase under single-hand support, the peak value of kinetic energy increased stepwise as the number of swings increased, and the mechanical energy remained higher than the target value.
On the other hand, when bar grasping was completed and the dual-hand support state became stable, the mechanical energy decayed and converged toward the target value.

Differences in motion characteristics were observed between the left-hand reaching motion and the right-hand reaching motion.
During left-hand motion, the swing period tended to be shorter and the amplitude larger, and the mechanical energy also showed relatively higher values.
In addition, the end-effector trajectory during left-hand motion exhibited greater variation compared with that during right-hand motion.

\textbf{Discussion}: 
In the hardware experiment, although the bars were arranged at equal intervals and constant height, the robot sometimes failed to grasp the target bar during the reaching motion.
One possible factor is approximation errors in the contact model.
In simulation, a collision model based on capsule approximation was adopted, which may lead to differences in contact characteristics compared with the hardware.
In particular, on hardware, the friction coefficient of the grasping surface was small and prone to slipping, which may have prevented sufficient support force from being generated during the wrap-around motion from above.

\begin{figure}[t]
  \centering
  \includegraphics[width=\columnwidth]{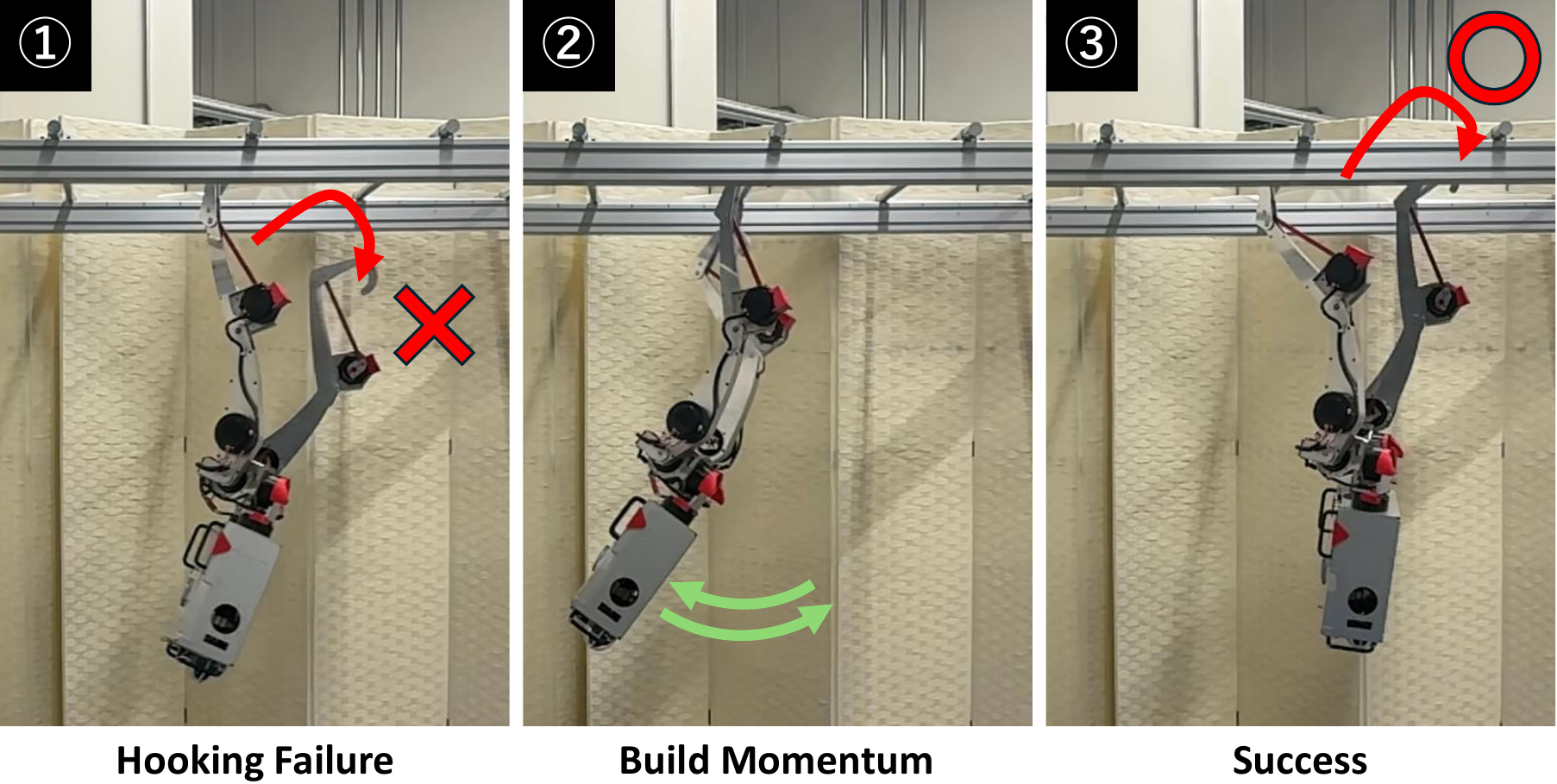}
  \vspace{-4ex}
  \caption{Failure recovery behavior during hardware brachiation. After an initial hooking failure, the robot increases swing amplitude to build momentum and successfully reattempts bar engagement, preventing fall and enabling continued progression.}
  \label{figure:retry}
  \vspace{-4ex}
\end{figure}

From the time-series variation of mechanical energy, the influence of the reward structure can be clearly observed.
Because the bar-reaching success reward is weighted larger than the energy-target tracking reward, the policy prioritizes forward progression during reaching even when the mechanical energy exceeds the target value.
In fact, during phases in which grasping failures were repeated, the peak value of kinetic energy increased stepwise as the number of swings increased.
As shown in \figref{retry}, this behavior can be interpreted as an amplification of the swing amplitude to extend the reachable distance of the end-effector, eventually leading to successful grasping.
On the other hand, once grasping is achieved and the state transitions to dual-hand support, the mechanical energy rapidly decays and converges toward the target value to obtain the energy reward.
This result indicates that the reward structure effectively switches the control objective between forward locomotion and stabilization.

Asymmetry between left-hand and right-hand motions was observed.
In the training environment, the initial motion phase of each episode was randomly selected between the left and right sides.
However, the learning logs indicate that the right-hand motion converged earlier than the left-hand motion.
This difference in acquisition speed likely resulted in differences in motion characteristics between the two sides on hardware, with the right-hand motion being more refined than the left-hand motion.
Since symmetry constraints were not explicitly introduced in this study, the asymmetry remained in the learned policy.

A particularly important observation in the hardware experiment was that failure recovery behavior consistently emerged.
The fact that the robot was able to restore its posture and continue swinging without falling after a grasping failure indicates that the brachiation motion strategy acquired by the proposed method remains effective even under the Sim-to-Real gap, including contact errors.

\section{CONCLUSIONS}

In this study, a learning method based on Waypoint-Guided Reinforcement Learning (WGRL) was proposed to acquire brachiation on a life-sized dual-arm robot.
WGRL enables guidance toward a target trajectory by specifying a small number of waypoints, even when full-body motion data for imitation learning are unavailable.
By combining a potential term that considers both distance and approach direction with rewards for progress and successful arrival, stable waypoint-following learning was achieved.
This resulted in the emergence of non-linear and complex motions for hook detachment and hooking.
Furthermore, the bar-reaching success reward promoting locomotion and the mechanical-energy tracking reward promoting motion stabilization were designed as paired elements.
Through this reward structure, a control strategy emerged in which the swing amplitude increases to reattempt grasping after failure, while the energy decreases after successful grasping to stabilize the motion.
As a result, robust forward progression was maintained even under geometric variations in the environment and in previously unseen courses.
Hardware validation also confirmed brachiation that sequentially transitions across bars without falling due to the emergence of failure recovery behavior.

This study demonstrates that RL-based brachiation can expand the traversable workspace of robots to environments without footholds.
Future work will focus on improving motion stability by introducing learning designs that explicitly promote symmetry between left and right motions, and on achieving more general arm-based locomotion by incorporating environment perception such as visual information.

\addtolength{\textheight}{-12cm}   


\bibliographystyle{junsrt}
\bibliography{main}

\end{document}